\documentclass{article}
\usepackage{natbib}
\usepackage[utf8]{inputenc} % allow utf-8 input
\usepackage[T1]{fontenc}    % use 8-bit T1 fonts
\usepackage{hyperref}       % hyperlinks
\usepackage{url}            % simple URL typesetting
\usepackage{booktabs}       % professional-quality tables
\usepackage{amsfonts}       % blackboard math symbols
\usepackage{nicefrac}       % compact symbols for 1/2, etc.
\usepackage{microtype}      % microtypography
\usepackage{lipsum}
\usepackage{graphicx}
\graphicspath{ {./images/} }
\usepackage{graphicx}
\usepackage{algorithm}
\usepackage{algpseudocode}
\usepackage{caption}
\usepackage{subcaption}

\title{Reinforcement Learning for Battery Management in Dairy Farming}

\author{
 Nawazish Ali \\
  School of Computer Science\\
  University of Galway\\
  Galway, Ireland H91 T8NW \\
  \texttt{n.ali3@universityofgalway.ie} \\
   \And
Abdul Wahid \\
  School of Computer Science\\
  University of Galway\\
  Galway, Ireland H91 T8NW \\
  \texttt{abdul.wahid@universityofgalway.ie} \\
  \And
 Rachael shaw \\
  Atlantic Technological University\\
  Galway, Ireland, H91 T8NW \\
  \texttt{rachael.shaw@atu.ie} \\
  \And
 Karl Mason \\
  School of Computer Science\\
  University of Galway\\
  Galway, Ireland H91 T8NW \\
  \texttt{karl.mason@universityofgalway.ie} \\
}

\begin{document}
{
    \renewcommand{\thefootnote}{\fnsymbol{footnote}}
    \footnotetext{\textit{Proc. of the Artificial Intelligence for Sustainability, ECAI 2023, Eunika et al. (eds.), Sep 30- Oct 1, 2023, https://sites.google.com/view/ai4s. 2023.}}
}
\maketitle
\begin{abstract}
Dairy farming is a particularly energy-intensive part of the agriculture sector. Effective battery management is essential for renewable integration within the agriculture sector. However, controlling battery charging/discharging is a difficult task due to electricity demand variability, stochasticity of renewable generation, and energy price fluctuations. Despite the potential benefits of applying Artificial Intelligence (AI) to renewable energy in the context of dairy farming, there has been limited research in this area. This research is a priority for Ireland as it strives to meet its governmental goals in energy and sustainability. This research paper utilizes Q-learning to learn an effective policy for charging and discharging a battery within a dairy farm setting. The results demonstrate that the developed policy significantly reduces electricity costs compared to the established baseline algorithm. These findings highlight the effectiveness of reinforcement learning for battery management within the dairy farming sector. 

\end{abstract}

% keywords can be removed

\section{Introduction}
The global population rise has increased food demand, including milk. Milk production rose from 735 to 855 million metric tons between 2000 and 2019. Growing demand for dairy products requires more electricity for farm activities. To reduce costs and reliance on external grids, adopting renewable energy like solar and wind is crucial \cite{b3,b4}.

Renewable energy generation has variability, so batteries are essential for storing electricity for future use. They play a crucial role in reducing dairy farming's electricity costs. To optimize battery performance, different techniques are used, including Maximizing Self-Consumption (MSC) and Time of Use (TOU) \cite{b5}. These methods involve regulating battery charge to reduce reliance on the power grid. With the advent of AI, many tasks have become more manageable.

AI has undergone significant advancements in recent years, following the emergence of the data revolution. Its potential has been demonstrated across various domains, offering encouraging outcomes. Reinforcement learning (RL) is a prevalent area of study within the field of artificial intelligence. RL agents are utilized in this domain and are not provided with explicit instructions on how to make decisions. Rather, they acquire knowledge through a process of trial and error by defining objective functions. Actor-Critical Learning and Q-Learning are widely recognized as the two most prominent Reinforcement Learning algorithms. RL agents have demonstrated efficacy in formulating policies across various domains. The objective of this paper is to apply RL agents to establish policies for battery management with the aim of improving renewable utilization and reducing electricity cost.

% In recent years, AI has advanced significantly due to the data revolution, showing promise across domains. Reinforcement Learning (RL), a key AI area, involves agents learning decisions through trial and error using objectives. Notably, Actor-Critical Learning and Q-Learning are leading RL algorithms. RL's efficacy in policy creation is evident. This paper aims to employ RL for battery management, enhancing renewable use and cutting electricity costs.

Previous studies have shown RL's effectiveness for residential battery management. Ebell et al. achieved a 7.8\% reduction in grid energy consumption using RL techniques with PV panels \cite{b8}. However, Minnaert et al. assessed energy storage options in dairy farming without utilizing RL algorithms \cite{b9}. RL's application to agriculture, especially in dairy farming, is limited in the existing literature. This study introduces RL for optimizing PV battery management in dairy farming and makes the following contributions:

\begin{enumerate}
  \item Development dairy farm energy simulator as a baseline.
  \item The implementation of Q-learning to control a battery store in a dairy farm.
  \item Compare the performance of Q-learning with existing baseline battery control algorithms. 
\end{enumerate}

\section{Background}
Battery management is a significant area of research across various domains, aiming to decrease reliance on power grid imports and cut costs. While many battery load regulation methods have been studied in scientific research, dairy farm battery management has not yet explored the use of RL. This study aims to implement battery management techniques in dairy farming to reduce power grid load. The main battery scheduling methods used in scientific studies include dynamic programming, rule-based scheduling, and Reinforcement Learning.

\subsection{Conventional Battery Control Methods}
Numerous investigations have explored operational strategies for PV Battery systems, focusing on objectives such as efficient utilization \cite{b13}. The MSC strategy, widely used for PV battery systems, particularly in distributed PV systems, efficiently consumes PV generation \cite{b14}. This approach aims to consume PV generation at the highest permissible rate, enhancing self-consumption \cite{b13}. Braun et al. found that proper battery utilization significantly increases local consumption of PV generation \cite{b15}. Researchers have optimized PV size \cite{b16} and storage capacity \cite{b17} to maximize self-consumption and minimize grid energy supply. Luthander et al.'s study indicates that appropriately sizing batteries can enhance relative self-consumption by 13-24% \cite{b18}.

Sharma et al. optimized battery size for zero-net energy homes with PV panels using the MSC strategy, potentially boosting self-consumption by 20-50\% \cite{b19,b20}. FiT and TOU tariffs promote PVB systems and demand-side engagement \cite{b21}. Christoph M Flath and Li et al. explored TOU tariff optimization methodologies \cite{b22,b23}. Prosumers seek economic benefits through FiTs and TOU strategies \cite{b24}. Gitizadeh et al. and Hassan et al. optimized battery capacity with tariff incentives \cite{b25,b26}. Ratnam et al. noted significant cost savings for PVB system users through FiTs \cite{b27}.

\subsection{Reinforcement learning for Battery Control}

Various applications utilize reinforcement learning algorithms to enhance energy management and efficiency. Wei et al. employed dual iterative Q-learning for smart residential battery management, optimizing charging and discharging for improved energy usage\cite{b28}. Kim et al. developed a reinforcement learning-based algorithm for smart energy-building management, dynamically determining optimal energy strategies\cite{b29}. Ruelens et al. applied reinforcement learning to enhance electric water heater efficiency by adapting to real-time demand and grid conditions\cite{b30}. Li et al. used multi-grid reinforcement learning to optimize HVAC systems for energy efficiency and comfort\cite{b31}.

Foruzan et al. proposed reinforcement learning for microgrid energy management, adapting to changing requirements and renewable generation\cite{b32}. Guan et al. suggested a reinforcement learning-based solution for domestic energy storage control, reducing costs through optimized charging and discharging\cite{b33}. Liu et al. employed deep reinforcement learning for household energy management, continuously learning optimal techniques for efficiency\cite{b34}. These applications demonstrate how reinforcement learning can enhance energy consumption and management efficiency across various contexts.

\section{Methodology}
\subsection{System Design}

The grid-connected photovoltaic (PVB) system includes solar PV, battery (Tesla Powerwall 2.0 with 13.5KWh capacity and 3.3 kW to 5 kW charging/discharging), power grid, and Dairy Farm load. PV power can go to the load, battery, or grid. The charge/discharge controller controls battery processes. The power grid powers the load, battery, PV, and battery storage as needed.

\subsection{Data and Price Profile}
The investigation used Load data from Finland \cite{b10} and PV data from the System Advisor Model (SAM) \cite{b35}. Hourly electricity usage was tracked for one year for a dairy farm with 180 cows, consuming 261 MWh annually. Similar-scale farms provided data, and a daily consumption profile was created from a large dairy farm's winter consumption. Sliding data series were made using monthly consumption data to estimate annual variations. Pricing data from the same area had hourly divisions into reduced, standard, and peak-hour rates: 11 pm to 7 am (reduced), 8 am to 4 pm and 7 pm to 10 pm (standard), and 5 pm to 7 pm (peak-hour).

\begin{figure}
\centerline{\includegraphics[width=2.8in, keepaspectratio]{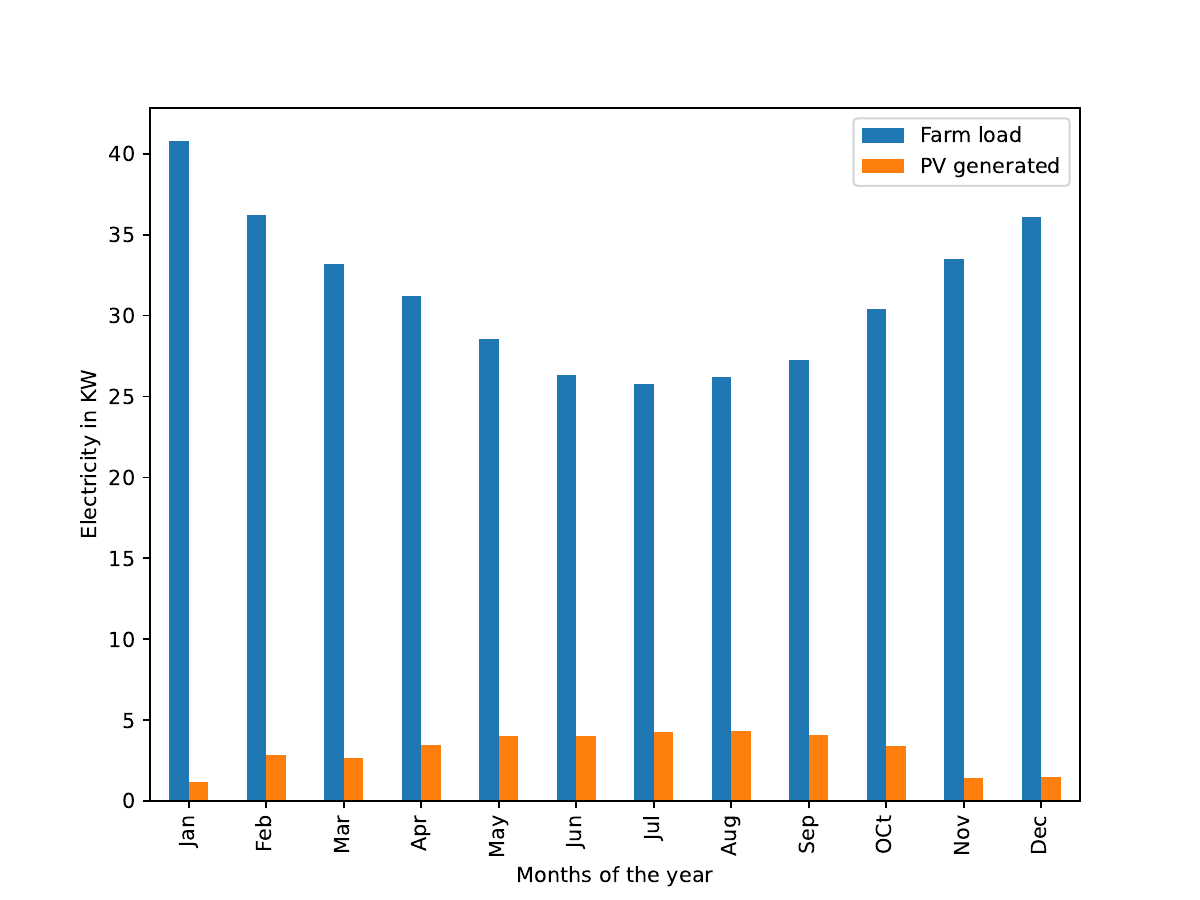}}
\caption{Farm load and PV generation for one year.} \label{load_data}
\end{figure}

\subsection{Baseline Implementation}
Two baseline algorithms are implemented which are MSC and TOU:

MSC optimizes PV power for load demand and battery charging in PV-integrated systems. The grid receives excess energy from the battery. When PV generation is low, the battery discharges to meet demand. Grid electricity is bought if demand exceeds PV and battery capacity. Weather drives MSC strategy.

TOU seeks economic gains from peak and off-peak electricity rates. Low-cost utility electricity charges the battery and discharges during high prices. Off-peak hours (23:00–8:00) charge the battery at the highest grid rate. Peak demand (5:00–7:00 PM) depletes the battery.

% \subsection{Reinforcement Learning Overview}
% Reinforcement Learning (RL) maximises cumulative reward from actions by interacting between an agent and its environment. RL is an MDP with the following components:
% State space: (\textbf{S}\)
% (\textbf{A}\) Action space
% Dynamic distribution: \(p(s_{t+1}|s_t; a_t)\)
% Reward function: R_t = sum_{i=t}^{T} \gamma^{(i-t)} r(S_i, A_i)\

% In each time step, the agent's behaviour is modified by the current state and prior actions and observations. \(\pi: S \rightarrow P(A)\) determines the agent's behaviour.

% R_t = \sum_{i=t}^{T} \gamma^{(i-t)} r(S_i, A_i)\), where \(\gamma\) is the discounting factor in the range [0, 1].

% RL seeks a policy \(\pi\) that maximises the initial probability distribution's expected total reward to maximise environmental reward.

\subsection{Application of Q-Learning to Battery Management}
The proposed methodology involves the utilization of the Q-learning Reinforcement Learning algorithm for efficient battery management. The optimal policy can be derived from the greatest Q values through Q-learning. The Q-Learning algorithm operates by choosing the action that corresponds to the maximum Q-value in every state. Equation \ref{max_q_value} illustrates the maximum Q value selection.

\begin{equation}
Q^*(s_t, a_t) = {argmax}_{a \in A} Q^\pi(s_t, a_t)
\label{max_q_value}
\end{equation}

The state space (\(\textbf{S}\)) influences an agent's perception and behavior. State space knowledge helps the RL agent optimize reward. State space complexity affects RL algorithm performance. Time and battery charge are considered in this study.

\subsubsection{Action Space (A)}
(A = \{0,1,2\}) The study considers charging (0), discharging (1), and remaining idle (2). The battery is charged by photovoltaic and utility grid electricity. To meet energy needs, action \(a = 1\) discharges the battery. Purchase grid power if the PV system and battery cannot provide enough energy. The dairy farm is idle and powered by solar PV and the grid (a = 2).

\subsubsection{Reward (\(R_p\))}

To reduce grid electricity export and import, this study optimizes PV system-generated electricity. The reward function, \(P_r\), is the grid electricity price minus 1.

\section{Results and Discussion}

Q-learning optimizes PV battery scheduling for solar power utilization and grid import reduction using reinforcement learning. Based on battery charge level, time of day, and energy prices, the Q-learning algorithm optimizes charging, discharging, and idle times over the year.

\begin{figure}[h]
     \centering
     \begin{subfigure}{0.49\textwidth}
         \centering
         \includegraphics[width=\textwidth]{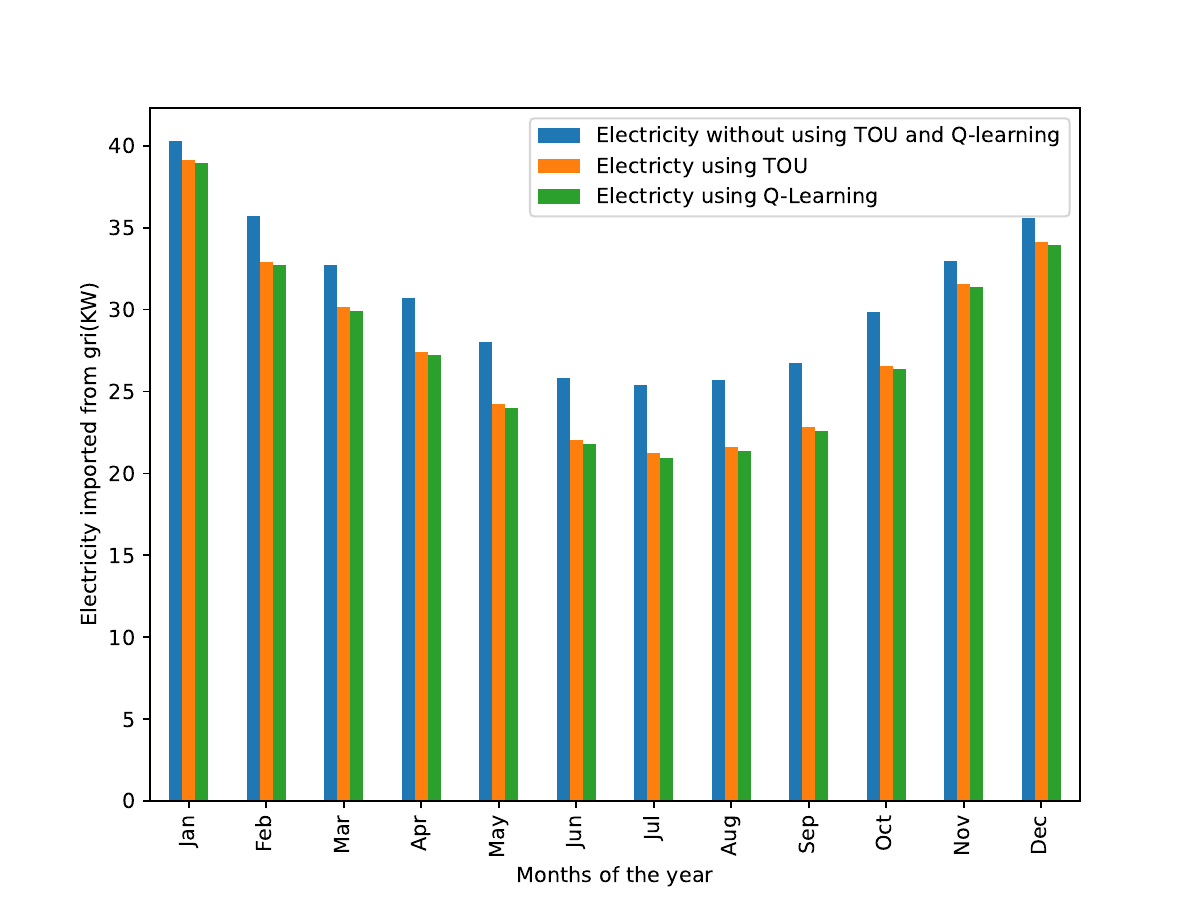}
         \caption{Electricity load comparison}
         \label{load_results}
     \end{subfigure}
     \hfill
     \begin{subfigure}{0.49\textwidth}
         \centering
         \includegraphics[width=\textwidth]{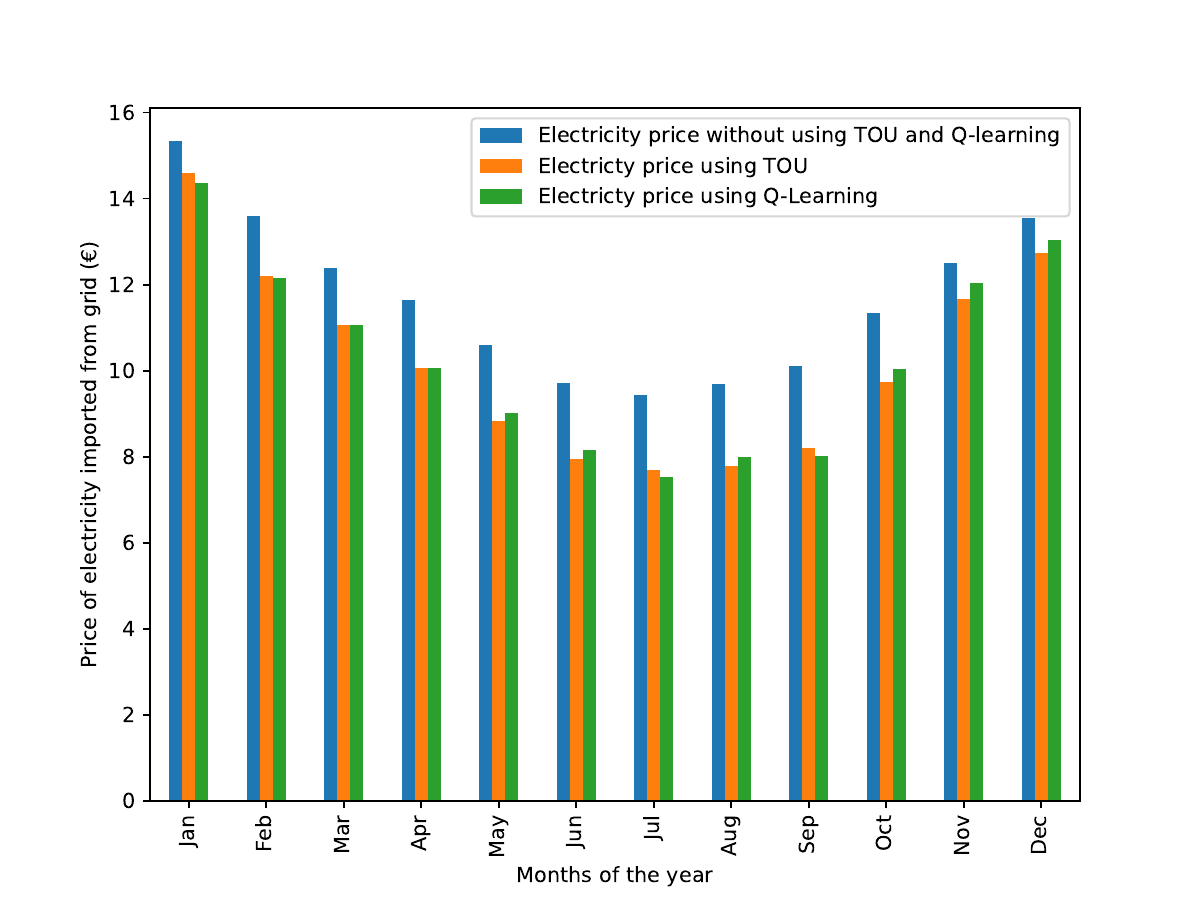}
         \caption{Electricity price comparison}
         \label{price_results}
     \end{subfigure}
        \caption{Comparison of the electricity load and price imported from the power grid by using Rule base and Q-learning.}
         \label{overall_comparsion}
\end{figure}

The Q-learning algorithm reduced the grid electricity importation from 9.45\% to 10.42\% as compared to the Rule base algorithm over the period of one year. As depicted in the figure \ref{load_results} Q-learning reduced a significant amount of energy importation from the external grid. Furthermore, the proposed algorithm reduces the cost of electricity imported from the grid by 11.93\% to 12.39\% compared to the rule base as illustrated in the figure \ref{price_results}.

Figure \ref{overall_comparsion} compares electricity consumption and importation over one year using various methods. Grid electricity purchased is on the vertical axis and hour on the horizontal. The graph shows three bars: grid-imported electricity without proposed algorithms, TOU-procured electricity, and Q-learning-consumed electricity. Electricity management affects grid importation and consumption.

\begin{figure}[h]
\centerline{\includegraphics[width=2.8in, keepaspectratio]{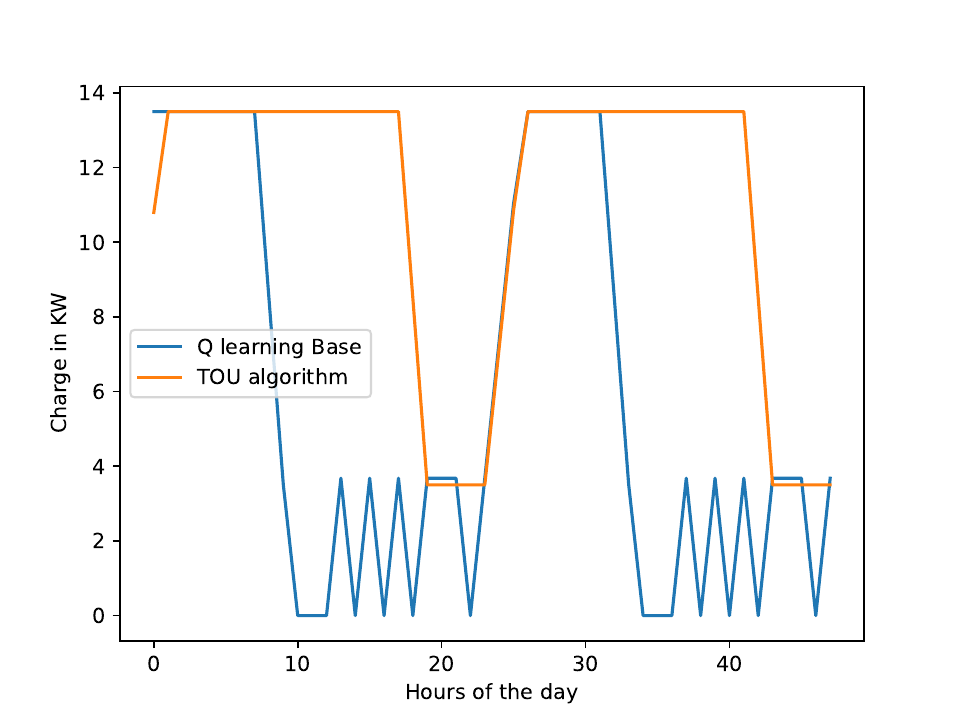}}
\caption{Comparison of the battery charging and discharging by using TOU and Q-learning.}
\label{battery_results}
\end{figure}

Figure \ref{battery_results} shows charging behavior over two days using various algorithms. The vertical axis shows battery charge in kilowatts, while the horizontal shows hours. Q-learning and rule-based battery charge lines. Battery charging affects system performance. Q-learning reduced electricity prices and load. The Q-learning agent charges the battery at night. The agent discharges the battery during peak prices or farm load demand to reduce grid electricity import. 

Q-learning maximizes dairy farm energy use. AI algorithms optimize PV battery scheduling to reduce electricity consumption without control. Dairy farms save energy with this.

\section{Conclusion}
\begin{enumerate}
  \item This paper discussed dairy farming battery load management. Two rule base algorithms in a dairy farm energy simulator reduced grid-imported electricity significantly. The rule base algorithm cuts imported electricity prices by 11.93\%. 
  \item This study shows that RL algorithms improve solar photovoltaic battery performance. RL can make real-time decisions in a continuous action space. RL beats TOU. Q-learning cuts electricity prices by 11.93–12.39%. 
  \item RL algorithms for dairy farm battery management may lower electricity prices because they outperform TOU algorithms. RL can be improved to handle complex and large systems.
  \item The Q-learning algorithm's battery management scalability for larger dairy farms could be studied. The Q-learning algorithm may be affected by weather and battery life. The study could compare Q-learning to other reinforcement learning algorithms, including deep reinforcement learning, to find the best dairy farm battery management method. Multiple batteries and better management could also be studied.

\end{enumerate}

\section*{Acknowledgements} This publication has emanated from research conducted with the financial support of Science Foundation Ireland under Grant number [21/FFP-A/9040].

\bibliographystyle{unsrt}  

\bibliography{template}

\begin{thebibliography}{10}

\bibitem{b3}
John Upton, Murphy Michael, Padraig French, Pat Dillon, et~al.
\newblock {\em Dairy farm energy consumption}.
\newblock 2010.
\newblock [Online; Accessed 20-April-2023].

\bibitem{b4}
Renewable energy opportunities for dairy farmers, 2021.
\newblock [Online; Accessed 20-April-2023].

\bibitem{b5}
Bin Zou, Jinqing Peng, Sihui Li, Yi~Li, Jinyue Yan, and Hongxing Yang.
\newblock Comparative study of the dynamic programming-based and rule-based
  operation strategies for grid-connected pv-battery systems of office
  buildings.
\newblock {\em Applied Energy}, 305:117875, 2022.

\bibitem{b8}
Niklas Ebell, Felix Heinrich, Jonas Schlund, and Marco Pruckner.
\newblock Reinforcement learning control algorithm for a pv-battery-system
  providing frequency containment reserve power.
\newblock In {\em 2018 IEEE International Conference on Communications,
  Control, and Computing Technologies for Smart Grids (SmartGridComm)}, pages
  1--6. IEEE, 2018.

\bibitem{b9}
Ben Minnaert, Bart Thoen, David Plets, Wout Joseph, and Nobby Stevens.
\newblock Optimal energy storage solution for an inductively powered system for
  dairy cows.
\newblock In {\em 2017 IEEE Wireless Power Transfer Conference (WPTC)}, pages
  1--4. IEEE, 2017.

\bibitem{b13}
Donald Azuatalam, Kaveh Paridari, Yiju Ma, Markus F{\"o}rstl, Archie~C Chapman,
  and Gregor Verbi{\v{c}}.
\newblock Energy management of small-scale pv-battery systems: A systematic
  review considering practical implementation, computational requirements,
  quality of input data and battery degradation.
\newblock {\em Renewable and Sustainable Energy Reviews}, 112:555--570, 2019.

\bibitem{b14}
Yijie Zhang, Tao Ma, Pietro~Elia Campana, Yohei Yamaguchi, and Yanjun Dai.
\newblock A techno-economic sizing method for grid-connected household
  photovoltaic battery systems.
\newblock {\em Applied Energy}, 269:115106, 2020.

\bibitem{b15}
Martin Braun, Kathrin B{\"u}denbender, Dirk Magnor, and Andreas Jossen.
\newblock Photovoltaic self-consumption in germany: using lithium-ion storage
  to increase self-consumed photovoltaic energy.
\newblock In {\em 24th European photovoltaic solar energy conference (PVSEC),
  Hamburg, Germany}, 2009.

\bibitem{b16}
DL~Talavera, FJ~Mu{\~n}oz-Rodriguez, G~Jimenez-Castillo, and C~Rus-Casas.
\newblock A new approach to sizing the photovoltaic generator in
  self-consumption systems based on cost--competitiveness, maximizing direct
  self-consumption.
\newblock {\em Renewable energy}, 130:1021--1035, 2019.

\bibitem{b17}
Neil~J Vickers.
\newblock Animal communication: when i’m calling you, will you answer too?
\newblock {\em Current biology}, 27(14):R713--R715, 2017.

\bibitem{b19}
Vanika Sharma, Mohammed~H Haque, and Syed~Mahfuzul Aziz.
\newblock Energy cost minimization for net zero energy homes through optimal
  sizing of battery storage system.
\newblock {\em Renewable Energy}, 141:278--286, 2019.

\bibitem{b20}
Emil Nyholm, Joel Goop, Mikael Odenberger, and Filip Johnsson.
\newblock Solar photovoltaic-battery systems in swedish
  households--self-consumption and self-sufficiency.
\newblock {\em Applied energy}, 183:148--159, 2016.

\bibitem{b21}
Luigi Dusonchet and Enrico Telaretti.
\newblock Comparative economic analysis of support policies for solar pv in the
  most representative eu countries.
\newblock {\em Renewable and Sustainable Energy Reviews}, 42:986--998, 2015.

\bibitem{b22}
Christoph~M Flath.
\newblock An optimization approach for the design of time-of-use rates.
\newblock In {\em IECON 2013-39th Annual Conference of the IEEE Industrial
  Electronics Society}, pages 4727--4732. IEEE, 2013.

\bibitem{b23}
Ran Li, Zhimin Wang, Chenghong Gu, Furong Li, and Hao Wu.
\newblock A novel time-of-use tariff design based on gaussian mixture model.
\newblock {\em Applied energy}, 162:1530--1536, 2016.

\bibitem{b24}
Na{\"\i}m~R Darghouth, Ryan~H Wiser, and Galen Barbose.
\newblock Customer economics of residential photovoltaic systems: Sensitivities
  to changes in wholesale market design and rate structures.
\newblock {\em Renewable and Sustainable Energy Reviews}, 54:1459--1469, 2016.

\bibitem{b25}
Mohsen Gitizadeh and Hamid Fakharzadegan.
\newblock Battery capacity determination with respect to optimized energy
  dispatch schedule in grid-connected photovoltaic (pv) systems.
\newblock {\em Energy}, 65:665--674, 2014.

\bibitem{b26}
Abubakar~Sani Hassan, Liana Cipcigan, and Nick Jenkins.
\newblock Optimal battery storage operation for pv systems with tariff
  incentives.
\newblock {\em Applied Energy}, 203:422--441, 2017.

\bibitem{b27}
Elizabeth~L Ratnam, Steven~R Weller, and Christopher~M Kellett.
\newblock An optimization-based approach to scheduling residential battery
  storage with solar pv: Assessing customer benefit.
\newblock {\em Renewable Energy}, 75:123--134, 2015.

\bibitem{b28}
Qinglai Wei, Derong Liu, and Guang Shi.
\newblock A novel dual iterative q-learning method for optimal battery
  management in smart residential environments.
\newblock {\em IEEE Transactions on Industrial Electronics}, 62(4):2509--2518,
  2014.

\bibitem{b29}
Sunyong Kim and Hyuk Lim.
\newblock Reinforcement learning based energy management algorithm for smart
  energy buildings.
\newblock {\em Energies}, 11(8):2010, 2018.

\bibitem{b30}
Frederik Ruelens, Bert~J Claessens, Salman Quaiyum, Bart De~Schutter,
  R~Babu{\v{s}}ka, and Ronnie Belmans.
\newblock Reinforcement learning applied to an electric water heater: From
  theory to practice.
\newblock {\em IEEE Transactions on Smart Grid}, 9(4):3792--3800, 2016.

\bibitem{b31}
Bocheng Li and Li~Xia.
\newblock A multi-grid reinforcement learning method for energy conservation
  and comfort of hvac in buildings.
\newblock In {\em 2015 IEEE International Conference on Automation Science and
  Engineering (CASE)}, pages 444--449. IEEE, 2015.

\bibitem{b32}
Elham Foruzan, Leen-Kiat Soh, and Sohrab Asgarpoor.
\newblock Reinforcement learning approach for optimal distributed energy
  management in a microgrid.
\newblock {\em IEEE Transactions on Power Systems}, 33(5):5749--5758, 2018.

\bibitem{b33}
Chenxiao Guan, Yanzhi Wang, Xue Lin, Shahin Nazarian, and Massoud Pedram.
\newblock Reinforcement learning-based control of residential energy storage
  systems for electric bill minimization.
\newblock In {\em 2015 12th Annual IEEE Consumer Communications and Networking
  Conference (CCNC)}, pages 637--642. IEEE, 2015.

\bibitem{b34}
Yuankun Liu, Dongxia Zhang, and Hoay~Beng Gooi.
\newblock Optimization strategy based on deep reinforcement learning for home
  energy management.
\newblock {\em CSEE Journal of Power and Energy Systems}, 6(3):572--582, 2020.

\bibitem{b10}
Sanna Uski and Erkka Rinne.
\newblock Data for a dairy farm microgrid solution, Jun 2018.

\bibitem{b35}
National Renewable Energy~Lab (NREL).
\newblock System advisor model (sam).
\newblock \url{https://sam.nrel.gov}, 2017.
\newblock [Online; Accessed 1-November-2022].

\end{thebibliography}

\end{document}